# A Tableau Calculus for Pronoun Resolution

Christof Monz[1,2] and Maarten de Rijke[2]

[1] Institute for Computational Linguistics (IMS), University of Stuttgart, Azenbergstr. 12, 70174 Stuttgart, Germany. E-mail: christof@ims.uni-stuttgart.de
[2] ILLC, University of Amsterdam, Plantage Muidergracht 24, 1018 TV Amsterdam, The Netherlands. E-mail: {christof, mdr}@wins.uva.nl

**Abstract.** We present a tableau calculus for reasoning in fragments of natural language. We focus on the problem of pronoun resolution and the way in which it complicates automated theorem proving for natural language processing. A method for explicitly manipulating contextual information during deduction is proposed, where pronouns are resolved against this context during deduction. As a result, pronoun resolution and deduction can be interleaved in such a way that pronouns are only resolved if this is licensed by a deduction rule; this helps us to avoid the combinatorial complexity of total pronoun disambiguation.

## 1 Introduction

The general aim of Natural Language Processing (NLP) is to analyze and understand human language using computational tools. In computational semantics, one of the subdisciplines of NLP, two specific tasks arise. First, what is the semantic value, the meaning, of a natural language utterance and how can we determine it. And, second, given the semantics of a natural language utterance, how can we use it to deduce further information? In practice, these questions are interdependent: to properly represent an utterance, one has to access contextual information and check what can be derived from it, and to perform derivations in the first place we obviously need to represent our information.

It is probably fair to say that developing inference methods for natural language is one of the most pressing tasks in computational semantics, and the present paper tries to contribute to this area. More specifically, we develop a tableau calculus in which deduction and pronoun resolution are interleaved. Before diving into the details in later sections, let us give a simple example of the natural language phenomenon that we are focusing on. Briefly, we are dealing with so-called *anaphoric expressions* or *anaphora*; typical examples of anaphora are pronouns such as 'she', 'he,' or 'it'. Anaphora are *resolved to* or *identified with* other terms, usually occurring earlier on in an utterance or discourse; such terms are called *antecedents*. Here's an example:

(1) A woman found a cat on a playground. She liked it.

What should 'it' in the second sentence in (1) refer back to—'a cat' or 'a playground'? As a rule, we, the human language users, don't have a problems resolving such ambiguities; in the case at hand 'a cat' would probably be selected as antecedent for 'it.' But how can a theorem prover that receives (1) as one of its premises use it to derive conclusions? As long as the pronoun 'it' has not been resolved, this question introduces the problem of ambiguity to the task of deduction with natural language semantics. Can we conclude either of the following from (1)?

(2) a. $\models$ A woman liked a cat.
    b. $\models$ A woman liked a playground.[1]

One way to tackle this problem is by assuming that anaphoric expressions have to be resolved before deductive methods are applied. This assumption is common in several approaches to the semantics of natural language, but in practice it seems to be too strong and highly implausible, since resolution of pronouns may in fact require deductive processing to be completed successfully [All94]. Here, we propose a different approach. We interleave disambiguation and deduction steps, where a pronoun is resolved only if this is needed by a deduction rule, and where deductive information is used to steer the resolution process.

In this paper, we assume that the semantic representations for natural language sentences are already given. Of course, this is not a trivial task, and it would be far beyond the scope of this paper to discuss this. [Als92] gives an overview of the Core Language Engine, an implementation builds (underspecified) semantic representations for natural language discourses.

The rest of the paper is organized as follows. In Section 2 we provide further examples and some linguistic background; this section may be skipped by anyone familiar with pronoun resolution. Then, in Section 3 we briefly introduce our formal language, and formalize the notion of context that we will need to model pronoun resolution interleaved with deduction. In Section 4 the semantics of our formal language is defined, and in Section 5 we provide it with a tableau calculus. Finally, Section 6 summarizes our results, and in Section 7 we draw some conclusions and formulate further challenges.

## 2  Some Linguistic Background

In this section we quickly review some basic facts and intuitions from natural language semantics as they pertain to pronoun resolution. Refer to [KR93] for further details.

If a sentence contains a pronoun, the hearer has to identify it with some person or thing that has been mentioned earlier to understand this sentence. Roughly, one can identify context with what has been said earlier. Of course, this blends out other contextual information like world knowledge, gestures, etc., but as these non-linguistic sources of context are hard to formalize, in general, we will restrict ourselves to the notion of context as linguistic context.

Saying that a pronoun has to be resolved to something that appears in the context does not mean that it can be identified with just anything in the context: there are some clear constraints. To illustrate these, we give some examples (the asterisk (*), indicates that a discourse is not well-formed).

The discourse in (3) below is not well-formed, because the pronoun 'She' and 'a man', which is the only thing that could function as an antecedent, do not agree on gender.

---

[1] To keep things simple, we do not employ any preference order of the readings, although this may be desirable in the long run.

(3) * A man sleeps. She snores.

In (4), it is not possible to bind 'it' to 'a car', because although 'a car' was mentioned before, its existence has not been claimed, on the contrary, it was said that there is no such car. In other words, 'it' cannot refer to something which is not existing.

(4) * Buk doesn't have a car. It is red.

Conditionals are another interesting case. The if-clause in (5.a) introduces two antecedents 'a linguist' and 'a car' which can both serve as antecedents for pronouns in the then-clause. But they cannot serve as antecedents for pronouns occurring in later sentences as in (5.b). Roughly, objects that are introduced in an if-clause are just assumed to exist, and the then-clauses expresses what has to hold, under this assumption. Clauses that follow the conditional sentence are not uttered within the context of this assumption, and therefore, their pronouns cannot access things occurring inside the assumption. One can say, that the assumption expressed by the if-clause is a local context, which is only accessible to the then-clause.

(5) a. If a linguist has a car, then it didn't cost much.
    b. * If a linguist has a car, then it didn't cost much. It is very old.

Universal quantification, as in (6), does not talk about particular individuals and it is not possible to refer back to 'every poet' with the pronoun 'he'. The same holds for indefinite noun phrases that occur in the scope of the universal quantifier. As they depend on each instantiation of the universally quantified variable, it does not mean that there has to be a particular individual which can be referred to by a singular pronoun.

(6) Every poet who has published a book likes it. $\begin{cases} \text{*He is arrogant.} \\ \text{*But it is really bad.} \end{cases}$

Summarizing the important points of the above examples, pronouns need to agree with their antecedents a number of features, and some information within a discourse may be inaccessible to pronouns that occur later in the discourse. These two points will play an important role in our tableau calculus below.

## 3 Towards Context-Based Reasoning

This section provides a formal account of context and the way it is dealt with in deduction. It will become obvious that deduction with natural semantics is much more structure-sensitive than for instance classical first-order logics. Here, we restrict ourselves to those kinds of structural information that is needed in order to allow pronoun resolution, but see [MdR98a] for a more general overview on this topic.

### 3.1 Formalizing Context

In the preceding section we provided some intuitions about pronoun-antecedent relations and the role structural information played in this setting. We will now formalize the way in which contextual information flows within a discourse. As a first step we introduce the formal language that we will be using.

**Definition 1 (The Language $\mathcal{L}^{pro}$).** Assuming that $\varphi_1$ and $\varphi_2$ are in $\mathcal{L}^{pro}$, we say that $\varphi$ is in $\mathcal{L}^{pro}$, if:

$$\varphi ::= R(x_{1_g}, \ldots, x_{n_{g'n}}) \mid \neg\varphi \mid \varphi_1 \wedge \varphi_2 \mid \varphi_1 \vee \varphi_2 \mid \varphi_1 \to \varphi_2 \mid$$
$$\forall x_g \varphi_1 \mid \exists x_g \varphi_1 \mid ?x_g \varphi_1$$

where $g, g' \in \{he, she, it\}$.

Thus, besides the usual logical connectives, $\neg, \wedge, \to, \vee, \forall, \exists$, we introduce a new operator ? that binds pronoun variables.

Contrary to approaches like Dynamic Predicate Logic (DPL, [GS91]) it is not assumed that pronouns are already resolved when constructing the semantic representation of a discourse. Given a formula $\varphi$, we say that $[\varphi]$ is a function from subsets of *VAR* (the set of variables) to subsets of *VAR*, where the argument is the *input context* and the value denotes the *output context* or *context contribution*. The contextual contribution of a formula $\varphi$ is the set of variables that $\varphi$ adds to the input context.

**Definition 2 (Contextual Contribution).** The *contextual contribution* of a formula $\varphi$ in $\mathcal{L}^{pro}$, $[\varphi]$, is defined recursively, as specified below. There, $i$ is a subset of *VAR*. Note that $[\cdot]$ is partial, where $[\varphi](i)$ is undefined whenever $\varphi$ contains pronouns that cannot be resolved against $i$.

(i) $[R(x_{1_g}, \ldots, x_{n_{g'}})](i) = \emptyset$
(ii) $[\neg\varphi](i) = \emptyset$, if $[\varphi](i)$ is defined
(iii) $[\varphi \wedge \psi](i) = [\varphi](i) \cup [\psi](i \cup [\varphi](i))$
(iv) $[\varphi \to \psi](i) = \emptyset$, if $[\varphi](i)$ and $[\psi](i \cup [\varphi](i))$ are defined
(v) $[\varphi \vee \psi](i) = \emptyset$, if $[\varphi](i)$ and $[\psi](i)$ are defined
(vi) $[\exists x_g \varphi](i) = \{x_g\} \cup [\varphi](i \cup \{x_g\})$
(vii) $[\forall x_g \varphi] = \emptyset$, if $[\varphi](i \cup \{x_g\})$ is defined
(viii) $[?x_g \varphi](i) = [\varphi](i)$, if $\exists y_g \in i$

Here $g, g' \in \{she, he, it\}$, and $[\varphi](i)$ is undefined, i.e., there is no $o$ such that $[\varphi](i) = o$, if the condition on the right hand side is not fulfilled. Undefinedness is preserved by set union: if $[\varphi](i)$ is undefined, then $[\varphi](i) \cup i'$ is not defined either, for any input $i'$.

Let us briefly discuss the above definition. Atomic formulas do not add variables to the input context and the output context is the empty set. Negation behaves as a *barrier*. In (ii), the output context is $\emptyset$, no matter what the output context of the formula in the scope of the negation is. Conjunction is totally dynamic: things introduced in the first conjunct can serve as antecedents for the second conjunct as well as for any later formula, and the output of the first conjunct contributes to the input of the second conjunct while the output of the second conjunct contributes to the output of the whole formula. The existential quantifier in (vi) adds the variable that it binds to the input context of its scope, but unlike the universal quantifier, it also adds the variable to the output context. In (viii), the pronoun operator is treated. It assumes that there is a variable $y$ in the input context that agrees with $x$ on gender.

## 3.2 Contextual Information and Deduction

Definition 2 explains how context flows through a sequence of formulas. As deductive methods such as the tableau method manipulate the structure of formulas, we have to guarantee that the flow of contextual information is preserved by these manipulations. Our informal discussion below explains how we achieve this by introducing suitable labels on formulas; the formal details are postponed until Section 5.

**Threading Context.** To resolve pronouns during deduction, it is necessary to keep track of the context against which a particular pronoun can be resolved. The context is not a global parameter because it can change while processing a sequence of formulas. To implement this idea, formulas will be annotated with contextual information of the form $(i,o):\varphi$, where $i$ is the input context and $o$ is the output context.

**Structure Preservation.** One of the major differences between dynamic semantics and classical logics is the structural sensitivity of the former. As an example, whereas $\neg\neg\varphi$ is classically equivalent to $\varphi$, this does not hold in dynamic semantics, because the output contexts of the formulas $\neg\neg\varphi$ and $\varphi$ are not the same since negation functions as a kind of barrier. Consider (7):

$$(7) \quad \frac{(i,\emptyset):\varphi \to \psi}{(i,o):\neg\varphi \quad | \quad (i\cup o, o'):\psi} \ (\to)$$

Neglecting labels, (7) is the regular tableau expansion rule for implication. Compare this to the definition of the contextual contribution of the implication in Definition 2, where the input context of the consequent $\psi$ consists of the union of the input context of the formula as a whole, $\varphi \to \psi$, *and* the set of contribution of the antecedent $\varphi$. Now in (7), $[\neg\varphi](i)$ will always equal $\emptyset$, simply because negation functions as a barrier; as explained in Section 2. Therefore, the implication rule $(\to)$ has it was stated in (7) gives the wrong results. Of course, the problem is that a negation sign has been introduced by a tableau rule, which is a violation of one of the major principles of deduction methods for natural language semantics, viz. preservation of structure. But this can easily be remedied by using signed tableaux, where each formula is adorned with its polarity. Reconsidering the tableau expansion rule for implication, we have to distinguish two cases: implication under positive and implication under negative polarity.

$$(8) \quad \frac{(i,\emptyset,+):\varphi \to \psi}{(i,o,-):\varphi \quad | \quad (i\cup o, o',+):\psi} \ (+:\to) \qquad \frac{(i,\emptyset,-):\varphi \to \psi}{(i,o,+):\varphi \atop (i\cup o, o',-):\psi} \ (-:\to)$$

Now, we can clearly distinguish between the truth-functional and contextual behavior of negation. Note, that both rules in (8) thread the context in a similar fashion even though their truth-functional behavior is different.

The order in which we process sentences is important, as they may contain anaphoric expressions that are only meaningful if the context provides an appropriate antecedent. This is also mirrored in the tableau rules where the input context of some node depends on the output of another node. For instance, $(i\cup o, o', -):\psi$ depends on $(i,o,+):\varphi$.

Observe that dependency does not only hold between formulas on the same branch, but can also occur between formulas on different branches, as exemplified by the rule $(+ : \rightarrow)$.

**A Note on Unification.** In Definition 2 contexts are defined as sets of variables. Below we will be using a free-variable tableau method (cf. [Fit96]), and we have to think about the double role of variables in a deduction: they are carriers of a value and possible antecedent for pronouns. Recall that in free-variable tableaux, universally quantified variables are substituted by a free variable and existentially quantified variables are substituted by a skolem function that depends on the free variables of the existentially quantified formula.

Consider the following situation. In a tableau, there are two nodes of the form $(i, o, p) : \varphi(x)$ and $(\{x\} \cup i', o', p') : \psi$, where $p, p' \in \{+, -\}$ and $x$ is a free variable in $\varphi$. If $x$ is instantiated to a term $t$, then we have to substitute $t$ for $x$ in all formulas. But do we also have to substitute $t$ for $x$ in all context parameters? In our calculus presented in Section 5 below, the following solution is adopted: If $t$ unifies with $t'$, then $t$ and $t'$ denote the same entity in the model that we are implicitly building while constructing a tableau. If $t$ is a possible antecedent for a pronoun $z$, then $t'$ has to be a possible antecedent for $z$, too, since $t$ and $t'$ simply denote the same entity. Therefore, term substitution is applied to both formulas and contexts.

**Introducing Goodness.** Up to now, labels adorning formulas carry two kinds of information: contextual information $(i, o)$ and polarity information $(+, -)$. These parameters reflect the dynamic behavior of natural language utterances and the way in which context is threaded through a sequence of sentences. In addition, we have to account for a more general restriction on natural language utterances. In the set-up that we have so far, if a pronoun $x_g$ occurs in a context $i$, all variables that are members of $i$ and agree on gender with $x_g$ can serve as antecedents. Unfortunately, this is too liberal, as the following example shows.

(9) A man sees a friend of his. He doesn't see him or he is in a rush.

The pronoun 'he' in (9) cannot refer to 'a man'. Intuitively, this would seem violate some kind of consistency constraint. To put it differently, a pronoun $x_g$ cannot be resolved to an antecedent $y_g$ if they carry contradictory information. Following [vD98], we call this restriction on possible pronoun resolutions *goodness*. Observe that goodness is a special case of a more general pragmatic principle like Grice's *maxim of quality*, cf. [Gri89].

How do we implement the notion of goodness in our calculus? The premises and the conclusion themselves should be consistent, as we assume that native speakers do not utter inconsistent sentences. In our calculus we will implement this idea by making an explicit distinction between the (original) premises and (original) conclusions of a tableau proof; we will mark the former with p and the latter with c.[2]

---

[2] Readers familiar with abduction may find it helpful to compare this distinction to the one where, in abduction, one requires that explanations preserve the consistency of the premises; see [CMP93].

Summing up, then, the nodes in our tableau calculus will be labeled formulas of the form $(i,o,\rho,p) : \varphi$. Here, $i$ is the input context, $o$ is the context contribution of $\varphi$, $\rho \in \{\mathsf{p},\mathsf{c}\}$ indicates whether $\varphi$ occurs as part of the premises or the conclusion, and $p \in \{+,-\}$ carries the polarity of $\varphi$.

## 4 The Semantics of Pronoun Ambiguity

Before we introduce our tableau calculus for $\mathcal{L}^{pro}$, we present its semantics and a notion of entailment for it. Starting with the latter, there are various possibilities. Following our discussion in Section 2, we opt for an entailment relation $\models_a$ where $\psi$ follows from $\varphi_1, \ldots, \varphi_n$ if there is a disambiguation $\theta$ of $\varphi_1, \ldots, \varphi_n$ and a disambiguation $\theta'$ of $\psi$ such that $\theta(\varphi_1, \ldots, \varphi_n) \models \theta'(\psi)$. This choice might lead to overdefinedness of some formula $\varphi$, since it might be the case that $M,h,i \models \theta(\varphi)$, for some disambiguation $\theta$, but $M,h,i \not\models \theta'(\varphi)$, for another disambiguation $\theta'$.

To be able to deal with this, we distinguish between *verification* ($\models_a$) and *falsification* ($\dashv_a$). To motivate this distinction, compare the sentences in (10).

(10) a. It is not the case that he sleeps.
    b. He doesn't sleep.

Their semantic representations are formulas of the form $\neg ?x_g\, \varphi$ and $?x_g\, \neg\varphi$, respectively. Intuitively, (10.a) and (10.b) have the same meaning, therefore it should be the case that $\neg ?x_g\, \varphi$ and $?x_g\, \neg\varphi$ are logically equivalent. If we would try to set up a semantics for $\mathcal{L}^{pro}$ by simply using $\models_a$, we would not get the desired equivalence of $\neg ?x_g\, \varphi$ and $?x_g\, \neg\varphi$:

(11) a.   $M,h,i \models_a \neg ?x_g\, \varphi$   iff   $M,h,i \not\models_a ?x_g\, \varphi$
                                     iff   $M,h[x_g/h(y_g)],i \not\models_a \varphi$ for all $y_g \in i$

    b.   $M,h,i \models_a ?x_g\, \neg\varphi$   iff   $M,h[x_g/h(y_g)],i \models_a \neg\varphi$ for some $y_g \in i$
                                    iff   $M,h[x_g/h(y_g)],i \not\models_a \varphi$ for some $y_g \in i$

The problem is that a semantics built up by using $\models_a$ interprets the ?-operator as a quantifier, which it is not. On the other hand, if we distinguish between verification and falsification, we get the desired equivalence:

(12) a.   $M,h,i \models_a \neg ?x_g\, \varphi$   iff   $M,h,i \dashv_a ?x_g\, \varphi$
                                     iff   $M,h[x_g/h(y_g)],i \dashv_a \varphi$ for some $y_g \in i$

    b.   $M,h,i \models_a ?x_g\, \neg\varphi$   iff   $M,h[x_g/h(y_g)],i \models_a \neg\varphi$ for some $y_g \in i$
                                    iff   $M,h[x_g/h(y_g)],i \dashv_a \varphi$ for some $y_g \in i$

In the following definition verification and falsification for the other boolean connectives are defined.

**Definition 3 (Semantics of $\mathcal{L}^{pro}$).** Verification and falsification are defined, given a model $M$, a variable assignment $h$ and a context $i$. As usual, a model $M = \langle \mathcal{U}, I \rangle$ consists of two parts: a universe $\mathcal{U}$ and an interpretation $I$ of the non-logical constants. First, we are going to define the semantics of the terms of $\mathcal{L}^{pro}$. $x_g$ is a variable (possibly a pronoun) with gender $g$, $c_g$ is a constant with gender $g$, and $f_g$ is a function with gender $g$, where $g \in \{he, she, it\}$

(a) $[\![x_g]\!]^{M,h,i} = h(x_g)$
(b) $[\![c_g]\!]^{M,h,i} = I(c_g)$
(c) $[\![f_g(t_1\ldots t_n)]\!]^{M,h,i} = f_g^I([\![t_1]\!]^{M,h,i}\ldots [\![t_n]\!]^{M,h,i})$

For formulas, $\models_a$ and $=\!|_a$ can be defined recursively:

(i) $M,h,i \models_a R(t_1,\ldots,t_n)$ iff $\langle [\![t_1]\!]^{M,h,i},\ldots,[\![t_n]\!]^{M,h,i}\rangle \in I(R)$
$M,h,i =\!|_a R(t_1,\ldots,t_n)$ iff $\langle [\![t_1]\!]^{M,h,i},\ldots,[\![t_n]\!]^{M,h,i}\rangle \notin I(R)$
(ii) $M,h,i \models_a \neg\varphi$ iff $M,h,i =\!|_a \varphi$
$M,h,i =\!|_a \neg\varphi$ iff $M,h,i \models_a \varphi$
(iii) $M,h,i \models_a \varphi \wedge \psi$ iff $M,h,i \models_a \varphi$ and $M,h,i\cup[\varphi](i) \models_a \psi$
$M,h,i =\!|_a \varphi \wedge \psi$ iff $M,h,i =\!|_a \varphi$ or $M,h,i\cup[\varphi](i) =\!|_a \psi$
(iv) $M,h,i \models_a \varphi \to \psi$ iff $M,h,i =\!|_a \varphi$ or $M,h,i\cup[\varphi](i) \models_a \psi$
$M,h,i =\!|_a \varphi \to \psi$ iff $M,h,i \models_a \varphi$ and $M,h,i\cup[\varphi](i) =\!|_a \psi$
(v) $M,h,i \models_a \varphi \vee \psi$ iff $M,h,i \models_a \varphi$ or $M,h,i \models_a \psi$
$M,h,i =\!|_a \varphi \vee \psi$ iff $M,h,i =\!|_a \varphi$ and $M,h,i =\!|_a \psi$
(vi) $M,h,i \models_a \exists x_g \varphi$ iff $\exists d \in \mathcal{U}: M,h[x_g/d],i \models_a \varphi$
$M,h,i =\!|_a \exists x_g \varphi$ iff $\forall d \in \mathcal{U}: M,h[x_g/d],i =\!|_a \varphi$
(vii) $M,h,i \models_a \forall x_g \varphi$ iff $\forall d \in \mathcal{U}: M,h[x_g/d],i \models_a \varphi$
$M,h,i =\!|_a \forall x_g \varphi$ iff $\exists d \in \mathcal{U}: M,h[x_g/d],i =\!|_a \varphi$
(viii) $M,h,i \models_a ?x_g \varphi$ iff $\exists y_g \in i: M,h[x_g/h(y_g)],i \models_a \varphi$
$M,h,i =\!|_a ?x_g \varphi$ iff $\exists y_g \in i: M,h[x_g/h(y_g)],i =\!|_a \varphi$

Overdefinedness is induced by the ?-operator. It is possible that a formula containing a pronoun can be verified *and* falsified by a model $M$; i.e., $M,h,i \models_a ?x_g \varphi$ and $M,h,i =\!|_a ?x_g \varphi$. In addition, it might also happen that the semantic value of a formula $\varphi$ is undefined. This is also due to (viii), where undefinedness results if there is no accessible variable in the context to which the pronoun could be resolved. In this case it holds that $M,h,i \not\models_a ?x_g \varphi$ and $M,h,i \neq\!|_a ?x_g \varphi$.

Thus, the resulting logic is four-valued containing besides truth (1) and falsity (0), overdefinedness and underdefinedness. This can be illustrated by a Hasse diagram, as in (13).

(13) 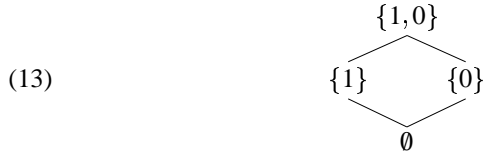

The sets denote the truth values that can be assigned to a formula. The singletons $\{1\}$ and $\{0\}$ denote the classical truth values, $\emptyset$ denotes undefinedness, and $\{1,0\}$ denotes overdefinedness.

Finally, we define a notion of entailment for sequences of formals that possibly contain unresolved pronouns.

**Definition 4 (Entailment in $\mathcal{L}^{pro}$).** Let $\varphi_1,\ldots,\varphi_n,\psi$ be in $\mathcal{L}^{pro}$, $h$ an arbitrary variable assignment, and $i$ an arbitrary context. We say that $\varphi_1,\ldots,\varphi_n$ *ambiguously entail* $\psi$, written as $\varphi_1,\ldots,\varphi_n \models_a \psi$, if for all $M$:

if for all $j \in \{1, \ldots, n\}$: $M, h, [\varphi_1 \wedge \cdots \wedge \varphi_{j-1}](i) \models_a \varphi_j$
then $M, h, i \models_a \psi$

Pronouns occurring in $\psi$ are resolved against the context $i$ which is also the context of the premises. Thereby, $\psi$ cannot pick antecedents introduced in the premises. Note, that there are several ways to define dynamic entailment relations and the one proposed is just one of them. [vB96, Chapter 7] classifies the entailment relation defined above as *update-to-update consequence*. Observe, by the way, that our notion of entailment is nonmonotonic, as most entailment relations in dynamic frameworks.

## 5  A Tableau Calculus for Pronoun Resolution

This section introduces our tableau calculus for reasoning with unresolved pronouns. The calculus consists of two components, a set of tableau expansion rules, and contextual parameters that allow us to interleave pronoun resolution and deduction steps. We first discuss the rules and then provide a short example.

### 5.1  The Tableau Expansion Rules

To reason with pronoun ambiguities we use a tableau calculus that is both free-variable and signed. The first property is simply to avoid the inefficiency of ground tableaux. Free-variable tableaux are fairly standard and we will not say much about them here; the reader is referred to [Fit96] instead. Signed tableaux are not new either, but here the signs are employed for a novel purpose. In Section 3, we motivated the distinction between negation in the object language ($\neg$) and polarities of tableau nodes ($+, -$). This was necessary because $\neg$ has an impact on the flow of contextual information, and to guarantee structure preservation we do not want to allow tableau rules to introduce additional negations. In addition, a distinction between verification and falsification is important to assign the right semantic values to formulas containing pronouns. An occurrence of a node of the form $+ : \varphi$ means that $\varphi$ is verifiable, which corresponds to $\models_a$, and an occurrence of the form $- : \varphi$ means that $\varphi$ is falsifiable, corresponding to $=\!\!|_a$.

The complete set of tableau rules constituting our calculus for pronoun ambiguity, $\mathcal{T}^{pro}$, is given in Table 1. The rules may seem somewhat overwhelming, but most of them are familiar ones. Remember that nodes are annotated by labels and are of the form $(i, o, \rho, p) : \varphi$, where $i$ is the input context, $o$ is the output context, which is computed by $[\varphi](i)$. $\rho$ indicates the origin of the formula, whether it occurred as a premise (p) or a conclusion (c). Polarity is simply expressed by $p$, $p \in \{+, -\}$. The way in which context is threaded through the tableau corresponds to the definition of contextual contribution, cf. Definition 2. Polarity assignment is done as defined in Definition 3.

Given our earlier discussions, the expansion rules should be obvious, but some rules deserve special attention. Let us first discuss the pronoun rules $(+ : ?)$ and $(- : ?)$. First, the ?-operator is simply dropped, and the variable it binds is substituted by one of its accessible terms that agrees with the pronoun on gender. These instantiations are marked as *pro*, in order to distinguish them from argument positions that are no instantiations of a pronoun. The set *PRO* is the set of all marked terms. The superscript has no influence

**Table 1.** The tableau rules for $\mathcal{T}^{pro}$

$$\frac{(i,o,\rho,+):\varphi\wedge\psi}{\begin{array}{c}(i,o',\rho,+):\varphi\\(i\cup o',o,\rho,+):\psi\end{array}}\;(+:\wedge)\qquad\frac{(i,o,\rho,-):\varphi\wedge\psi}{(i,o',\rho,-):\varphi\;\;\Big|\;\;(i\cup o',o,\rho,-):\psi}\;(-:\wedge)$$

$$\frac{(i,\emptyset,\rho,+):\varphi\vee\psi}{(i,o,\rho,+):\varphi\;\;\Big|\;\;(i,o',\rho,+):\psi}\;(+:\vee)\qquad\frac{(i,\emptyset,\rho,-):\varphi\vee\psi}{\begin{array}{c}(i,o,\rho,-):\varphi\\(i,o',\rho,-):\psi\end{array}}\;(-:\vee)$$

$$\frac{(i,\emptyset,\rho,+):\varphi\rightarrow\psi}{(i,o,\rho,-):\varphi\;\;\Big|\;\;(i\cup o,o',\rho,+):\psi}\;(+:\rightarrow)\qquad\frac{(i,\emptyset,\rho,-):\varphi\rightarrow\psi}{\begin{array}{c}(i,o,\rho,+):\varphi\\(i\cup o,o',\rho,-):\psi\end{array}}\;(-:\rightarrow)$$

$$\frac{(i,\emptyset,\rho,+):\neg\varphi}{(i,o,\rho,-):\varphi}\;(+:\neg)\qquad\frac{(i,\emptyset,\rho,-):\neg\varphi}{(i,o,\rho,+):\varphi}\;(-:\neg)$$

$$\frac{(i,\emptyset,\rho,+):\forall x_g\,\varphi}{(i\cup\{x_g\},o,\rho,+):\varphi[x_g/X_g]}\;(+:\forall)\qquad\frac{(i,\emptyset,\rho,-):\forall x_g\,\varphi}{(i\cup\{x_g\},o,\rho,-):\varphi[x_g/f_g(X_1\ldots X_n)]}\;(-:\forall)^{\dagger}$$

$$\frac{(i,o\cup\{x_g\},\rho,+):\exists x_g\,\varphi}{(i\cup\{x_g\},o,\rho,+):\varphi[x_g/f_g(X_1\ldots X_n)]}\;(+:\exists)^{\dagger}\qquad\frac{(i,o\cup\{x_g\},\rho,-):\exists x_g\,\varphi}{(i\cup\{x_g\},o,\rho,-):\varphi[x_g/X_g]}\;(-:\exists)$$

$$\frac{(i\cup\{t_g\},o,\rho,+):?x_g\,\varphi}{(i\cup\{t_g\},o,\rho,+):\varphi[x_g/t_g^{pro}]}\;(+:?)\qquad\frac{(i\cup\{t_g\},o,\rho,-):?x_g\,\varphi}{(i\cup\{t_g\},o,\rho,-):\varphi[x_g/t_g^{pro}]}\;(-:?)$$

$$\frac{\begin{array}{c}(i,o,\rho,+):R(s_1,\ldots,s_n)\\(i',o',\rho',-):R(t_1,\ldots,t_n)\end{array}}{\bot}\;(\bot)^{\ddagger}$$

$^{\dagger}$Where $X_1\ldots X_n$ are the free variables in $\varphi$ and $i$.
$^{\ddagger}$If $\rho\neq\rho'$ then $\{s_1,\ldots,s_n,t_1,\ldots,t_n\}\cap PRO=\emptyset$

on unification of terms, it is just needed to constrain the closure of a branch to cases that obey goodness.

Next, we consider the rules $(+ : \exists)$ and $(- : \forall)$. Both rules involve skolemization, and the question is which influence pronoun variables have on skolem terms. Consider the node in (14).

(14)   $(i, o, \rho, +) : \exists x_g \, ?y_{g'} \, \varphi$

In (14), applying the tableau expansion rule $(+ : \exists)$ will substitute $x_g$ by a skolem function $f_g(X_1 \ldots X_n)$, where $X_1, \ldots, X_n$ are the free variables in $\varphi$. But what about $y_{g'}$? It does not occur free in $\varphi$, because it is bound by the ?-operator, but it could be resolved to some $t_n$ in the context, which contains free variables. This dilemma is due to the order of application. First, skolemization is carried out, and then pronoun resolution. This leads to incorrectness. For instance, from $\forall x_g \exists y_{g'} \, ?z_g \, R(z_g, y_{g'})$ we can now derive $\exists y_{g'} \forall x_g \, ?z_g \, R(z_g, y_{g'})$, which is clearly not a valid derivation. Here, $x_g$ does not occur overtly in $R(z_g, y_{g'})$, but $z_g$ can be resolved to $x_g$. To fix this, skolemization does not only have to depend on the free variables occurring in formulas, but also on the free variables occurring in the terms of the input context since pronouns can be resolved to these terms.

Finally, $(\bot)$ carries the proviso that two literals of distinct polarity, where both originate from the premises (marked p) or both originate from the conclusion (marked c), do not allow to close a tableau branch if they contain pronoun instantiations. This allows us to encode goodness into the tableau calculus, saying that pronouns can only be resolved to antecedents that do not carry contradictory information, as exemplified by (9). It ensures that the premises themselves and the conclusion on its own are interpreted consistently. But of course, it is still possible to derive a contradiction from the combination of the former with the negation of the latter.

### 5.2 An Example

Given a two-sentence sequence *A man sees a boy. He whistles*, we want to see whether we can derive *A man whistles*. The semantic representation of the premises is given by the first two nodes, and the negation of the conclusion is given by the third node. The corresponding proof is displayed in Table 2.

First, we try to resolve the pronoun to $g_{he}$. This allows us to close the rightmost branch, with mgu $\{U_{he}/g_{he}\}$. But then there is no contradictory node for $(\{g_{he}\}, \emptyset, \text{c}, -) : man(g_{he})$. Hence, we apply pronoun resolution again, and this time resolve it to $f_{he}$. Next, universal instantiation is applied with the new free variable $V_{he}$. Now, all remaining branches can be closed by the mgu $\{V_{he}/f_{he}\}$. In Table 2, the pairs that allow to close a branch are connected by a dashed line.

The threading of contextual information may seem a bit confusing, but it is hard to display the dynamics of the instantiation of the context variables on 'static' paper. It may be helpful to read the tableau rules in Table 1 as PROLOG clauses, where the context variables of the parent of a rule unify with the context variables of the node the rule is applied to.

**Table 2.** A Sample Proof in $\mathcal{T}^{pro}$

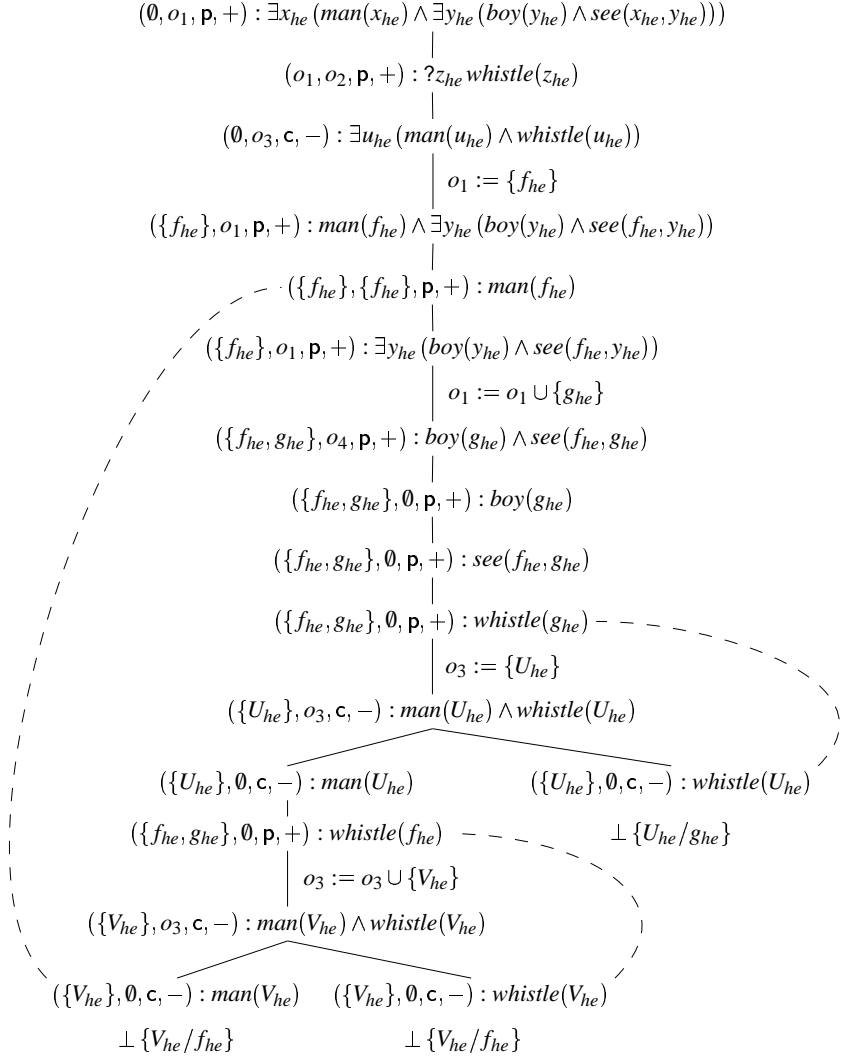

## 6 Results

The tableau calculus $\mathcal{T}^{pro}$ has a number of advantages over a resolution-based approach to pronoun resolution, as provided in [MdR98b]. First of all, it is possible to interleave the computation of accessible variables with deduction, since preservation of structure is guaranteed in our signed tableau method. This is not possible in resolution, because it is assumed that the input is in conjunctive normal form, which destroys all structural information needed for pronoun binding. Accessible antecedents can only be computed by a preprocessing step, cf. [MdR98b].

But the major advantage is that no backtracking is needed if the choice of an antecedent for a pronoun does not allow us to close all open branches; we simply apply pronoun resolution again, choosing a different antecedent. Of course, more has to be said about controlling proof construction than we have room for here. For instance, one would like to prevent the proof method from choosing again an antecedent for a pronoun that did not allow to close some branches. This can be accomplished by some simple book keeping about the antecedents that have been used before.

$\mathcal{T}^{pro}$ has been implemented in PROLOG, and is based on leanT$^A$P [BP95,PSar], a well-known depth-first theorem prover for classical first-order logic. It is slightly adapted for our purposes, where we dispense with the assumption that the input is in negation normal form as this violates the principle of structure preservation. Of course, this adaption results in a less lean, but still rather efficient theorem prover. The PROLOG implementation of $\mathcal{T}^{pro}$ is available online at $\sim$ .

To conclude this section, let us turn to a brief discussion of soundness and completeness of $\mathcal{T}^{pro}$. There are at least two strategies for establishing soundness and completeness. Of course, one can follow a *direct* strategy: prove soundness by in the traditional manner, and prove completeness by using the 'classical' completeness proof for free variable tableaux based on Hintikka sets is adapted. Here, we sketch an *indirect* strategy that consists in reducing the soundness and completeness of $\mathcal{T}^{pro}$ to the soundness and completeness of a traditional free-variable tableau calculi for first-order logic, $\mathcal{T}^{class}$. The basic intuition is the following: by analyzing tableaux for $\mathcal{T}^{pro}$ one can extract pronoun resolutions that can be used to help preprocess ambiguous $\mathcal{L}^{pro}$ formulas and turn them into traditional first-order formulas, while preserving enough information about satisfiability.

**Theorem 5 (Soundness of $\mathcal{T}^{pro}$).** *If a closed tableau can be generated from* $\Gamma = \{(\emptyset, o, \mathsf{p}, +) : \bigwedge_{k=1}^{n} \varphi_k, (\emptyset, o'', \mathsf{c}, -) : \psi\}$, *where* $\varphi_1, \ldots, \varphi_n, \psi \in \mathcal{L}^{pro}$, *then* $\varphi_1, \ldots, \varphi_n \models_a \psi$.

*Proof.* Given a closed tableau $T$ for $\Gamma$, the pronoun instantiations $\{t_1^{pro} \ldots t_m^{pro}\}$ that led to the closure of the branches of $T$ are collected. Then, we relate the instantiations to the pronoun variables $\{x_1 \ldots x_m\}$ that introduced them by an application of $(+ : ?)$ or $(- : ?)$. As $\{t_1 \ldots t_m\}$ are free variables or skolem terms, we identify the quantifier variables $\{y_1 \ldots y_m\}$ that introduced $\{t_1 \ldots t_m\}$. This yields two substitution of the form $\theta = \{x_1/y_1 \ldots x_j/y_j\}$ and $\theta' = \{x_{j+1}/y_{j+1} \ldots x_m/y_m\}$, where $\theta$ disambiguates the pronouns occurring in the premises, and $\theta'$ disambiguates the pronouns occurring in the conclusion. To ensure that substituted variables are classically bound, we apply a re-bracketing algorithm which is used in dynamic semantics, in order to relate dynamic semantics to classical logic, cf. [GS91]. To illustrate the process of re-bracketing, it allows us to replace a dynamic formula such as $\exists x \varphi(x) \wedge \psi(x)$ by its classical counterpart $\exists x (\varphi(x) \wedge \psi(x))$. More generally the re-bracketing algorithm may be specified as follows:

**Definition 6 (Re-bracketing).** Every dynamic formula can be translated to a formula of classical first-order logic. In [GS91] a function b is defined that accomplishes this. b is defined recursively:

1. $bR(t_1 \ldots t_n) = R(t_1 \ldots t_n)$
2. $b \neg \varphi = \neg b \varphi$
3. $b(\varphi_1 \vee \varphi_2) = b\varphi_1 \vee b\varphi_2$
4. $b \exists x \varphi = \exists x\, b \varphi$
5. $b \forall x \varphi = \forall x\, b \varphi$
6. $b(\varphi_1 \wedge \varphi_2) =$
    (a) $b(\psi_1 \wedge (\psi_2 \wedge \varphi_2))$ if $\varphi_1 = \psi_1 \wedge \psi_2$
    (b) $\exists x\, b(\psi \wedge \varphi_2)$ if $\varphi_1 = \exists x \psi$
    (c) $b\varphi_1 \wedge b\varphi_2$ otherwise
7. $b(\varphi_1 \to \varphi_2) =$
    (a) $b(\psi_1 \to (\psi_2 \to \varphi_2))$ if $\varphi_1 = \psi_1 \wedge \psi_2$
    (b) $\forall x\, b(\psi \to \varphi_2)$ if $\varphi_1 = \exists x \psi$
    (c) $b\varphi_1 \to b\varphi_2$ otherwise

Re-bracketing can be applied, because pronoun variables are always substituted by quantified variables that are accessible in the sense of Definition 2, which is based on the notion of accessibility in dynamic semantics, see e.g., [KR93,GS91].

Then, a closed tableau for $\varphi_1, \ldots, \varphi_n, \neg \psi$ in $\mathcal{T}^{pro}$ gives rise to a closed tableau for $\varphi'_1, \ldots, \varphi'_n, \neg \psi'$ in $\mathcal{T}^{class}$, by the soundness of $\mathcal{T}^{class}$, where $\varphi'_1, \ldots, \varphi'_n$ is the result of applying $\theta$ and re-bracketing to $\varphi_1, \ldots, \varphi_n$, and $\neg \psi'$ results from applying $\theta'$ and re-bracketing to $\neg \psi$ □

**Theorem 7 (Completeness of $\mathcal{T}^{pro}$).** *If an open tableau can be generated from* $\Gamma = \{(\emptyset, o, \mathsf{p}, +) : \bigwedge_{k=1}^{n} \varphi_k, (\emptyset, o'', \mathsf{c}, -) : \psi\}$*, where* $\varphi_1, \ldots, \varphi_n, \psi \in \mathcal{L}^{pro}$*, then* $\varphi_1, \ldots, \varphi_n \not\models_a \psi$.

*Proof.* If $\Gamma$ is consistent in $\mathcal{T}^{pro}$, then it may be shown that for all admissible pronoun resolutions $\theta, \theta'$, the set $\theta\varphi_1, \ldots, \theta\varphi_n, \theta' \neg \psi$ is consistent in $\mathcal{T}^{class}$. Obviously, we neeed to get rid of ambiguous formulas involving the ?-operator. when moving from $\mathcal{T}^{pro}$ to $\mathcal{T}^{class}$, but this is what the admissible pronoun resolution does for us. By the completeness of $\mathcal{T}^{class}$ (see [Fit96]), we get that $\theta\varphi_1, \ldots, \theta\varphi_n, \theta' \neg \psi$ is (classically) satisfiable, for any admissible $\theta, \theta'$. Hence, $\varphi_1, \ldots, \varphi_n, \neg \psi$ is satisfiable according to $\models_a$, as required. □

## 7 Conclusions

In this paper, we have proposed a tableau calculus that tries to tackle an instance of a particularly important and difficult task in computational semantics: automated reasoning with ambiguity. A tableau calculus that allows one to interleave disambiguation and deduction has been proposed to overcome the problem of state explosion one inevitably runs into if theorem proving is applied to naïvely disambiguated semantic representations.

To enable on-the-fly disambiguation during proof development, it is necessary that enough structural information of the original representation is preserved. In the case of pronoun resolution this structural information is needed to define which variables can serve as antecedents for pronouns. The nodes in the tableau were adorned with labels containing this additional contextual information.

It turned out that tableau methods are especially well-suited for reasoning with natural language semantics, since they are analytic (in contrast to Natural Deduction), and they allow for a more sensitive manipulation of the syntactic structures of the formulas (in contrast to resolution methods). See, for instance, [KK98,MdR98c] for other applications of tableau methods in the area of computational semantics.

Future work will be devoted to extending our tableau calculus to more complex cases of anaphora resolution, like presuppositions, or plural pronouns, where contextual information has to contain more structure than just lists of accessible terms. At the same time, it has to be investigated how a more comprehensive framework that allows to reason with different kinds of ambiguity can be set up. We plan to combine our tableau calculus for pronoun resolution with some of our earlier work on reasoning with quantificational ambiguity, cf. [MdR98c].

**Acknowledgments.** We want to thank the referees for their helful comments. Christof Monz was supported by the Physical Sciences Council with financial support from the Netherlands Organisation for Scientific Research (NWO), project 612-13-001. Maarten de Rijke was supported by the Spinoza Project 'Logic in Action' at ILLC, the University of Amsterdam.